\documentclass[runningheads]{llncs}
\usepackage[T1]{fontenc}
\usepackage{color}
\usepackage{url} 

\usepackage{cite}
\usepackage{amsmath,amssymb,amsfonts}
\usepackage{algorithmic}
\usepackage{graphicx}
\usepackage{textcomp}
\usepackage{xcolor}
\usepackage{float}
\usepackage{subfigure}
\usepackage{multirow}
\usepackage{algorithmic}
\usepackage[ruled,linesnumbered]{algorithm2e}
\begin{document}

\title{Discriminative Graph-level Anomaly Detection via Dual-students-teacher Model\\}

\titlerunning{GLADST}
\author{Fu Lin\inst{1,2}\and
Xuexiong Luo\inst{2}\and
Jia Wu\inst{2}\and
Jian Yang\inst{2}\and
Shan Xue\inst{2}\and
Zitong Wang\inst{1}\and
Haonan Gong\inst{1}}

\authorrunning{Fu Lin et al.}
\institute{Wuhan University, Wuhan, Hubei, China \and
Macquarie University, Sydney, NSW, Australia \\
\email{\{linfu, zitongwang, gonghaonan\}@whu.edu.cn}\\
\email{\{jia.wu, jian.yang, emma.xue\}@mq.edu.au,} \email{xuexiong.luo@hdr.mq.edu.au} \\
}

\maketitle

\begin{abstract}
Different from the current node-level anomaly detection task, the goal of graph-level anomaly detection is to find abnormal graphs that significantly differ from others in a graph set. Due to the scarcity of research on the work of graph-level anomaly detection, the detailed description of graph-level anomaly is insufficient. Furthermore, existing works focus on capturing anomalous graph information to learn better graph representations, but they ignore the importance of an effective anomaly score function for evaluating abnormal graphs. Thus, in this work, we first define anomalous graph information including node and graph property anomalies in a graph set and adopt node-level and graph-level information differences to identify them, respectively. Then, we introduce a discriminative graph-level anomaly detection framework with dual-students-teacher model, where the teacher model with a heuristic loss are trained to make graph representations more divergent. Then, two competing student models trained by normal and abnormal graphs respectively fit graph representations of the teacher model in terms of node-level and graph-level representation perspectives. Finally, we combine representation errors between two student models to discriminatively distinguish anomalous graphs. Extensive experiment analysis demonstrates that our method \footnote{The source code is at \url{ https://github.com/whb605/GLADST.git}} is effective for the graph-level anomaly detection task on graph datasets in the real world.
\keywords{graph anomaly detection  \and graph neural networks \and dual-students-teacher model}
\end{abstract}

\section{Introduction}
Graph anomaly detection investigation has already become a hot topic in academic and industry communities in the past few years. Researchers aim to design a more effective anomaly detection method to detect existing anomalous information on graph datasets \cite{ma2021comprehensive,yang2023comprehensive}. Besides, they also actively explore practical application scenarios based on graph anomaly detection task, such as abnormal account detection on financial transaction platforms \cite{zhang2022efraudcom}, fake information monitor on social websites \cite{shu2017fake,khattar2019mvae} and intrusion detection in cyber security \cite{zhou2021hierarchical}. However, most of the current research pays more attention to analyzing 
abnormal nodes from a graph, i.e., node-level anomaly detection. For example, DOMINANT \cite{ding2019deep}, ComGA \cite{luo2022comga} and DAGAD \cite{liu2022dagad} models utilize deep graph neural networks (GNNs) \cite{wu2020comprehensive} to capture various node anomalies including local, global and community structure anomalies in the single graph. Even though these methods have achieved great success, a new problem involving how to detect existing anomalous graphs within a set of graphs is worth further exploration, and it also has the huge practical value, such as distinguishing abnormal molecule graphs for molecule property prediction.

Due to the obvious difference between node-level anomaly detection and graph-level anomaly detection, the previous approaches are not appropriate for graph-level anomaly work. Thus, we initially need to explore the key problem, this being which form does an abnormal graph take compared with other normal graphs. According to the intuitive analysis, an abnormal graph will represent a significant difference in node and graph properties. Specifically, a certain node may contain anomalous attribute information and have an abnormal connection with neighbors. For example, when we monitor bank account transactions in a certain region, abnormal accounts will show the account identity information anomaly and many abnormal transaction connections with others. Furthermore, the graph property anomaly more shows the difference of the whole graph structure information. For example, the molecule graph with two benzene rings is abnormal compared with other molecule graphs that only have a benzene ring on a molecule graph set. Thus, two key anomaly definitions above are conducive to find out anomalous graphs within a graph set. Another key problem is to design an effective anomaly score function to judge which graph is abnormal. Besides, the score function has the power to discriminatively distinguish normal and abnormal graphs without the influence of graph data types. It is worth mentioning that there is a limited quantity of research about the graph-level anomaly detection problem, such as GLocalKD \cite{ma2022deep}, GLADC \cite{luo2022deep}, and iGAD \cite{zhang2022dual} methods. But these methods either ignore two types of graph anomaly form mentioned or they do not consider a discriminative anomaly score function for anomalous graph detection.

Based on the aforementioned discussion, in this article, we design a discriminative \textbf{G}raph \textbf{L}evel \textbf{A}nomaly \textbf{D}etection framework by building a competitive dual-\textbf{S}tudents-\textbf{T}eacher model named \textbf{GLADST}. The proposed GLADST framework consists of one teacher model, two student models and an anomaly score function, where the backbone of these models are GNNs. Specifically, we first train the teacher model with a heuristic loss to make learned graph representations more divergent, which can help better capture complex graph information pattern. Then we train student model A with normal graphs to fit the graph representation distribution of the teacher model from node-level and graph-level representation perspectives. Similarly, student model B is trained by abnormal graphs according to the way above. The key idea of the design is that two competing student models can effectively learn normal and abnormal graph representation patterns, respectively and node-level and graph-level representation errors achieve node and graph properties anomaly detection, respectively. Thus, given a test graph, if the graph is normal, graph representations of student model A will better match graph representations of the teacher model, and student model B will keep away. In other words, node-level and graph-level representation errors of student model A and the teacher model will be smaller, but for student model B, these will be larger. Besides, to discriminatively distinguish abnormal graphs and normal graphs, we design a competitive anomaly score function based on the representation error value of two student models, which makes the anomaly score of the real abnormal graph larger than the normal graph. Thus, our work has the following three key contributions:
\begin{itemize}
	\item We explore the graph-level anomaly detection problem and define existing anomalous graph information including node and graph property anomalies. Furthermore, we jointly utilize node-level representation and graph-level representation errors to detect these anomalies.
	\item We introduce a discriminative graph-level anomaly detection framework relying on dual-students-teacher model. Specifically, the special training way is beneficial to better learning normal and abnormal graph representation patterns. And the anomaly score function relies on the representation error value of two student models, it can optimize the effectiveness of abnormal graph detection.
	\item We conduct performance comparison experiments with baselines and model analysis experiments to illustrate the efficiency of GLADST.
\end{itemize}
\section{Related Work}
Benefiting from advanced deep learning techniques, graph anomaly detection research based on GNNs has attracted considerable interest recently. In light of the difference between anomalous objects, graph anomaly detection can be categorized into the two types below:

\textbf{Node-level Anomaly Detection} (NLAD) is to discover abnormal nodes which are different from other nodes in structure and attribute information. And NLAD is used to identify abnormal nodes by inputting a graph. DOMINANT \cite{ding2019deep} first employed GNNs for the NLAD task. It utilizes GNNs to learn graph representations and then constructs the reconstruction errors of graph structure and node attribute to capture abnormal nodes. Afterwards, many methods based on GNNs \cite{fan2020anomalydae,li2019specae,luo2022comga,liu2022dagad} focus on analyzing different types of node anomalies, such as local, global, and community structure anomalies in graphs. In addition, some NLAD methods based on graph contrastive learning are proposed \cite{liu2021anomaly,zheng2021generative,chen2022gccad,jin2021anemone} and they built different contrast pairs of node and subgraph to better exploit rich graph information for anomalous node detection.
    
\textbf{Graph-level Anomaly Detection} (GLAD) detects abnormal graphs that have the obvious difference with other graphs in a graph set. Besides, GLAD is clearly different from NLAD and the aforementioned methods are unsuitable for the GLAD task. Thus, several research works have explored this issue. For example, GLocalKD \cite{ma2022deep} utilized the predictor network to learn normal graph representations of the random network by global and local graph representation distillation and abnormal graphs will show obvious graph representation errors in the framework. But the method easily fails these graph data that abnormal graph representation pattern is not very obvious within the graph set. GLADC \cite{luo2022deep} used disturbed features to construct contrastive instances to improve the performance of GLAD. iGAD \cite{zhang2022dual} designed a new graph neural network to investigate anomalous attributes and substructures to learn graph representations. Other two methods \cite{liu2023good,ligraphde} focus more on the out-of-distribution problem of graph data. Although these methods achieve great performance in GLAD task, they lack an effective anomaly score function to keep competitive performance on different types of abnormal graph data. Thus, we propose a discriminative GLAD framework, where two competing student models learn normal and abnormal graph representation patterns respectively by a trained teacher model. Then the representation error value between two student models can be significantly distinguished abnormal and normal graphs.

In addition to the advancements in graph neural networks for anomaly detection, there are several notable developments in graph processing techniques that contribute to the field. For example, Hooi, B. et al. proposed a method\cite{hooi2016fraudar} to analyze the real-world graphs on fraud attacks. DenseAlert and DenseStream\cite{shin2017densealert} can focus on detecting dense subtensors to discover the anomalies. Spade\cite{jiang2022spade} is another method that proposed three fundamental peeling sequence reordering techniques. It can effectively detect fraudulent communities.

\section{Definition and Problem Statement}
\paragraph{Deﬁnition 1 (Graph)}
$G=\left(\mathcal{V}_{G},\mathcal{X}_{G},\mathcal{E}_{G},\mathcal{A}_{G}\right)$
represents a graph, where $\mathcal{V}_{G}=\{v_{1},v_{2},...,v_{n}\}$ denotes the node set and $x_{i}\in \mathcal{X}_{G}$ is the attribute feature of node $v_{i}\in \mathcal{V}_{G}$. $\mathcal{X}_{G}$ is the attribute feature matrix. We call the graph $G$ \emph{plain graph} if it doesn't have the attribute information, otherwise, it is called \emph{attributed graph}. $e_{ij}\in \mathcal{E}_{G}$ is the edge between $v_{i}$ and $v_{j}$. $\mathcal{A}_{G}$ is the adjacency matrix, $\mathcal{A}_{G}(i,j)=1$ denotes that nodes $v_{i}$ and $v_{j}$ have an edge between them; and $\mathcal{A}_{G}(i,j)=0$ otherwise. 
\paragraph{Deﬁnition 2 (Graph-level Anomaly)} 
Given a graph dataset $\mathcal{G}=\{G_{1},G_{2},...,G_{m}\}$ with each graph $G\in \mathcal{G}$ denoted by $G=\left(\mathcal{V}_{G},\mathcal{X}_{G},\mathcal{E}_{G},\mathcal{A}_{G}\right)$. \textbf{Node property anomaly} is where the node of given graph $G$ has anomalous attributes and abnormal connections with neighbors compared with normal graphs. \textbf{Graph property anomaly} is when the structure construction of graph $G$ is inconsistent with others in $\mathcal{G}$ from the global view.

We aim to learn an anomaly evaluation function $f:\mathcal{G}\to \mathbb{R}$ with parameter $\Theta$ on the graph set $\mathcal{G}$, and the return value of function $f(\hat{G}_{i};\Theta)>f(\hat{G}_{j};\Theta)$ when the input graph $\hat{G}_{i}$  is more like an anomaly graph than $\hat{G}_{j}$. 
\section{Framework of GLADST}
\begin{figure*}[b!]
	\small
	\includegraphics[width=\textwidth]{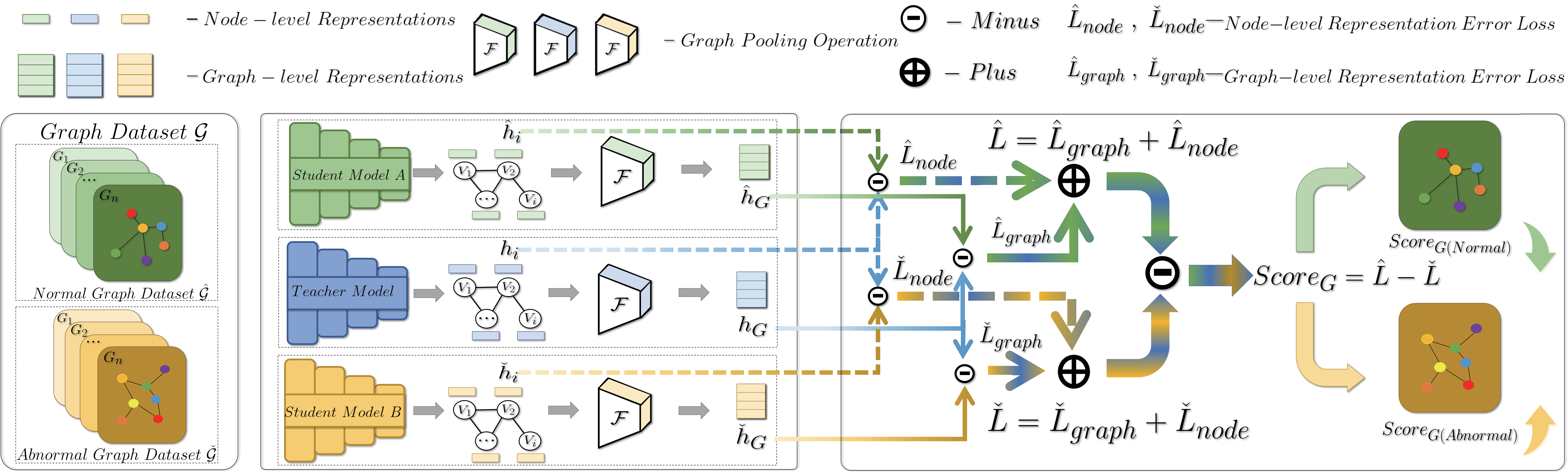}
	\caption{The framework of the proposed GLADST. We first train the teacher model with a heuristic loss to learn node-level and graph-level representations on a given graph dataset. Then, we train the student model A to obtain normal node-level and graph-level representation patterns respectively by the teacher model under node-level and graph-level representation error loss optimization. Similarly, the student model B is trained to obtain abnormal node-level and graph-level representation patterns. And the backbone of these models are GNNs. Finally, the value of representation error between student A and student B is as the anomaly score to identify anomalous graphs.} \label{fig1}
\end{figure*}
To capture normal and abnormal graph representation patterns respectively and learn an effective anomaly score function, we propose a discriminative graph-level anomaly detection framework. As shown in Figure \ref{fig1}, the framework is composed of a dual-students-teacher model and a discriminative anomaly score function, and the detailed operation is introduced as follows:

\subsection{Dual-students-teacher Model}
As GLocalKD framework \cite{ma2022deep} used a predictor network to capture normal graph representation pattern of the random network by the knowledge distillation method, but this way easily gets suboptimal performance when abnormal graph pattern is difficult to be distinguished. We consider to design a dual-students-teacher model to overcome the above problem.
\subsubsection{Trained Teacher Model.}\label{SCM} The teacher model is a graph convolutional network (GCN) \cite{kipf2016semi} that aggregates node's neighbors feature information to update itself feature, to learn graph representations. The teacher model takes matrix $\mathcal{A}_{G}$ and $\mathcal{X}_{G}$ as input and then uses GCN to map each node $v_{i}\in \mathcal{V}_{G}$ into the representation space. We define $h_{i}^{l}$ as the hidden representation of node $v_{i}$ at the $l^{th}$ layer :
\begin{equation}
h_{i}^{l} = ReLU\left(\mathcal{\tilde{D}}_{G}^{-1/2}\mathcal{\tilde{A}}_{G}\mathcal{\tilde{D}}_{G}^{-1/2}h_{i}^{l-1}\Theta^{l-1}\right), \label{eq8}
\end{equation}
where the $\left(l-1\right)^{th}$ layer's weight parameters are $\Theta^{l-1}$ and the node representation is $h_{i}^{l-1}$. $\mathcal{\tilde{A}}_{G}=\mathcal{A}_{G}+\mathcal{I}_{G}$ and $\mathcal{I}_{G}$ denotes the identity matrix. $\left|G\right|$ is the number of nodes and $\mathcal{\tilde{D}}_{G}$ is the corresponding diagonal degree matrix:
\begin{equation}
\mathcal{\tilde{D}}_{G}\left ( i,i \right )=\sum_{j=1}^{\left|G\right|}\mathcal{\tilde{A}}_{G}\left ( i,j \right ),\label{eq9}
\end{equation}
where the feature vector in $\mathcal{X}_{G}$ is used as the initialized input of node representations, \emph{i.e}, the $0^{th}$ layer's $h_{i}^{0}=\mathcal{X}_{G}(i,:)$. The plain graph $G$ does not have the parameter $\mathcal{X}_{G}$, following \cite{zhang2018end,ma2022deep}, so we construct a simple $\mathcal{X}_{G}$ by using the node degree information. 

The $h_{i}$ of the last layer is the model's output of node-level representations. We use the max-pooling operation for all node representations on $d$ dimension space, and learn graph-level representations $h_{G}$ :  
\begin{equation}
h_{G}=[\max_{i=1}^{n}h_{i,1},\max_{i=1}^{n}h_{i,2},...,\max_{i=1}^{n}h_{i,d}]. \label{eq10}
\end{equation}

We utilize a dataset comprising graphs $\mathcal{G}$ to train the teacher model and initialize the model weights $\Theta$ randomly. Specifically, we develop a heuristic loss to form the teacher model and the training purpose is to make learned graph representations more divergent, which can help student models better capture normal and abnormal graph patterns. The training losses are as:
\begin{equation}  
L_{teacher}=\frac{1}{\left ( L_{graph} + L_{node} \right )}, \label{eq11}
\end{equation}
\begin{equation} 
L_{graph} = \frac{1}{\left|\mathcal{G}\right|}\sum_{G\in\mathcal{G}}^{} Std\left( h_{G}\right), \label{eq12}
\end{equation}
\begin{equation}
L_{node} = \frac{1}{\left|\mathcal{G}\right|}\sum_{G\in\mathcal{G}}^{} \left (  \frac{1}{\left|G\right|}\sum_{v_{i}\in\mathcal{V}_{G}}^{} Std\left ( h_{i}\right ) \right ), \label{eq13}
\end{equation}
where $Std\left (.\right )$ is the standard deviation function where a high standard deviation suggests that the values are spread out. When the loss $L_{teacher}$ is minimized, $L_{graph}$ and $L_{node}$ become larger, \emph{i.e}, graph-level representations $h_{G}$ and node-level representations $h_{i}$ are spread out over a wider range.

\subsubsection{Double Student Models.} To capture normal and abnormal graph representation patterns respectively based on the trained teacher model, we design two competing student models that are indispensable in the training process and their backbones are GCN model \cite{kipf2016semi} with exactly the same structure as the teacher model. Then, we will describe the work flow of two student models. 
\begin{enumerate}
    \item We initially input the normal graph dataset $\hat{\mathcal{G}} \in \mathcal{G}$ into the trained teacher model above to acquire node-level representations $h_{i}$ and graph-level representations $h_{G}$. Then, we also train the student model A based on graph set $\hat{\mathcal{G}}$ to learn node-level representations $\hat{h}_{i}$ and graph-level representations $\hat{h}_{G}$. 
    \item We construct node-level representation error loss $\hat{L}_{node}$ and graph-level representation error loss $\hat{L}_{graph}$ with the trained teacher model, which aims to catch normal node-level and graph-level representation patterns. The two losses as:
    \begin{equation}
\hat{L}_{graph} =  \frac{1}{\left|\mathcal{G}\right|}\sum_{G\in\mathcal{G}}^{} f_{d}\left ( h_{G},\hat{h}_{G} \right ), \label{eq3}
    \end{equation}
    \begin{equation}
\hat{L}_{node} =   \frac{1}{\left|\mathcal{G}\right|}\sum_{G\in\mathcal{G}}^{} \left (  \frac{1}{\left|G\right|}\sum_{v_{i}\in \mathcal{V}_{G}}^{} f_{d}\left ( h_{i},\hat{h}_{i} \right ) \right ),        \label{eq4}
    \end{equation}
    where $f_{d}(\cdot,\cdot)$ is the function to calculate the difference between two graph representations. Here, we can choose \emph{mean square error (MSE)} function.
    \item We can learn a final normal graph representation pattern by the following loss as:
    \begin{equation}
\hat{L} =\hat{L}_{graph} + \hat{L}_{node}. \label{eq1}
    \end{equation}
    \item Finally, we use abnormal graph dataset $\check{\mathcal{G}} \in \mathcal{G}$ to train the student model B, and the training process of student B is the same as that of model A as described above. And we also can learn a final abnormal graph representation pattern $\check{L}$.
\end{enumerate}

\subsection{Discriminative Anomaly Score Function}
In our framework, we consider node-level representation error and graph-level representation error to detect two categories of graph anomaly respectively: \emph{node property anomaly} and \emph{graph property anomaly}. Specifically, we propose a dual-students-teacher model above to capture normal and abnormal graph representation patterns. Thus, when we input a test graph sample $G$, the discriminative anomaly score function is designed as follows:
\begin{equation}
\begin{split}
Score_{G}=\left (\left\| h_{G}-\hat{h}_{G}\right\|^{2} +  \frac{1}{\left|G\right|}\sum_{v_{i}\in \mathcal{V}_{G}}^{}  \left\| h_{i}-\hat{h}_{i} \right\|^{2}\right ) \\ - \left(\left\| h_{G}-\check{h}_{G}\right\|^{2} +  \frac{1}{\left|G\right|}\sum_{v_{i}\in \mathcal{V}_{G}}^{}  \left\| h_{i}-\check{h}_{i} \right\|^{2} \right ). \label{eq14}
\end{split}
\end{equation}

If the value of $Score_{G}$ is larger, the probability that graph $G$ is an abnormal graph is greater.
\subsection{Theoretical Analysis}
We use model $\phi$ to represent the teacher model, model $\hat{\phi}$ to represent student model A and model $\check{\phi}$ to represent student model B. Given a test graph sample $G \in \mathcal{G}$, $\phi^{*}_{G}$ denotes the representations
outputs of the teacher model, $\hat{\phi}^{*}_{G}$ and $\check{\phi}^{*}_{G}$ denote the representation outputs of these two student models, respectively. The score of anomaly is simplified as follows:
\begin{equation}
Score_{G}=\hat{S}_{G}-\check{S}_{G}=\left\|\phi^{*}_{G}-\hat{\phi}^{*}_{G}\right\|^{2}- \left\|\phi^{*}_{G}-\check{\phi}^{*}_{G}\right\|^{2}.  \label{eq15}
\end{equation}

At the training stage, we first use $\mathcal{G}$ to train the
teacher model $\phi$ with a heuristic loss. Then we use normal graphs $\hat{\mathcal{G}}$ to train student model $\hat{\phi}$ and use abnormal graphs $\check{\mathcal{G}}$ to train student model $\check{\phi}$. We want node-level representations and graph-level representations of two student models on each training sample to be as close as possible to the corresponding representations of the teacher model, respectively. Thus, other training graphs with similar patterns will have small prediction errors between them in the student model A. The situation is similar for the student model B. Specifically, given a normal graph sample $G$, its patterns are similar to many other training graphs of normal graph set $\hat{\mathcal{G}}$, and the loss error $\hat{S}_{G}$ is small, by contrast, $\check{S}_{G}$ is large because its patterns are dissimilar to many other training graphs of abnormal graph set $\check{\mathcal{G}}$. Thus, if $G$ is normal, $\hat{S}_{G}$ is small and $\check{S}_{G}$ is large, after normalization, $Score_{G}=\hat{S}_{G}-\check{S}_{G}<0$ under ideal conditions. Otherwise, if $G$ is abnormal, $\hat{S}_{G}$ is large and $\check{S}_{G}$ is small, after normalization, the anomaly score $Score_{G}=\hat{S}_{G}-\check{S}_{G}>0$ under ideal conditions. Obviously, the anomaly score above can be significantly distinguished normal and abnormal graphs compared with current baselines whose anomaly scores only rely on simple graph representation errors. 

\section{Experiments}
\subsection{Datasets}
\begin{table}[b!]
\caption{The information of experimental datasets.}
\centering

\begin{tabular}{c|ccc}
\hline
Datasets & Graphs & Avg-nodes & Avg-edges \\ \hline
HSE      & 8,417   & 16.89     & 17.23     \\
MMP      & 7,558   & 17.62     & 17.98     \\
P53      & 8,903   & 17.92     & 18.34     \\
PPAR     & 8,451   & 17.38     & 17.72     \\ \hline
AIDS     & 2,000   & 15.69     & 16.20     \\
BZR      & 405    & 35.75     & 38.36     \\
COX2     & 467    & 41.22     & 43.45     \\
DHFR     & 756    & 42.43     & 44.54     \\
NCI1     & 4,110   & 29.87     & 32.30     \\ \hline
ENZYMES  & 600    & 32.63     & 62.14     \\
PROTEINS & 1,113   & 39.06     & 72.82     \\ \hline
COLLAB   & 5,000   & 74.49     & 2,457.78   \\ \hline
\end{tabular}

\label{statistics_tab}
\end{table}

\begin{table*}[h!]
\caption{Anomaly detection performance measured mean value of AUC (\%) and standard deviation (\%) when graph data of label $0$ is graph anomaly.}
\centering
\begin{tabular}{ccccccc}
\hline
Datasets & FGSD-IF             & FGSD-LOF   & FGSD-OCSVM & GLocalKD   & GOOD-D              & GLADST              \\ \hline
HSE      & 39.38\tiny±1.35          & 43.44\tiny±2.49 & 42.24\tiny±4.43 & 59.25\tiny±1.09 & \textbf{69.39\tiny±1.05} & 54.76\tiny±2.12          \\
MMP      & 67.78\tiny±0.90          & 57.00\tiny±1.98 & 52.14\tiny±2.97 & 32.43\tiny±0.81 & \textbf{69.76\tiny±8.10} & 68.50\tiny±0.72          \\
P53      & 66.94\tiny±3.67          & 56.55\tiny±3.55 & 48.63\tiny±2.76 & 33.35\tiny±3.34 & 62.51\tiny±1.85          & \textbf{68.86\tiny±3.51} \\
PPAR     & 34.49\tiny±4.03          & 46.41\tiny±4.81 & 50.45\tiny±3.86 & 65.46\tiny±4.05 & \textbf{66.65\tiny±1.47} & 61.75\tiny±3.12          \\ \hline
AIDS     & \textbf{99.38\tiny±0.89} & 87.73\tiny±4.89 & 86.20\tiny±4.22 & 96.61\tiny±0.53 & 92.58\tiny±1.36          & 97.65\tiny±0.98          \\
BZR      & 44.51\tiny±6.06          & 49.56\tiny±8.73 & 41.15\tiny±6.44 & 67.12\tiny±8.71 & 74.84\tiny±5.40          & \textbf{81.60\tiny±2.80} \\
COX2     & 56.49\tiny±3.45          & 56.71\tiny±4.85 & 54.23\tiny±5.81 & 52.13\tiny±7.24 & 61.17\tiny±7.49          & \textbf{63.35\tiny±7.44} \\
DHFR     & 51.62\tiny±5.25          & 49.20\tiny±5.94 & 55.89\tiny±4.45 & 63.11\tiny±3.38 & 61.17\tiny±4.82          & \textbf{76.67\tiny±2.63} \\
NCI1     & 33.19\tiny±1.59          & 53.93\tiny±2.18 & 50.18\tiny±2.58 & 68.32\tiny±1.47 & 60.32\tiny±2.39          & \textbf{68.44\tiny±0.81} \\ \hline
ENZYMES  & 48.51\tiny±5.96          & 38.98\tiny±6.57 & 42.80\tiny±8.79 & 55.27\tiny±1.40 & 63.10\tiny±4.29          & \textbf{71.77\tiny±5.84} \\
PROTEINS & 75.40\tiny±2.79          & 59.79\tiny±3.64 & 33.63\tiny±1.64 & 68.55\tiny±5.31 & 72.18\tiny±3.96          & \textbf{79.60\tiny±3.93} \\ \hline
COLLAB   & 45.42\tiny±1.49          & 61.47\tiny±1.27 & 37.55\tiny±1.26 & 51.95\tiny±1.36 & \textbf{70.55\tiny±2.15} & 52.76\tiny±1.52          \\ \hline
\end{tabular}
\label{tab1}
\end{table*}

We perform experiments to showcase the efficiency and adaptability of the model we proposed on diverse datasets. Therefore, we choose twelve public and available real-world datasets and their statistics are given in Table \ref{statistics_tab}. HSE, MMP, p53, PPAR are real graph anomalies. They are chemical compounds with complex and different structures in toxicology studies and the unique structure may make the compound activity different in certain conditions. Furthermore, these datasets have been categorized into test and training sets in original setting and here we mix them up and redivide them in our experiment. AIDS, BZR, COX2, DHFR and NCI1 are molecule datasets, where every node symbolizes an atom in the molecule, and every edge symbolizes a chemical bond. ENZYMES and PROTEINS are protein datasets. The difference is that nodes here mean amino acids, and edges indicate that the connected nodes are relatively close. COLLAB is a social network dataset. The nodes are individuals, and the connections are edges. Thus, the coverage of experimental datasets is wide enough to examine the capability of our model. Besides, the degree information of the node is chosen as the node attribute feature for these plain graph data, according to \cite{zhang2018end,ma2022deep}. It is worth noting that all these datasets including real graph anomaly and classification graph data are bifurcated into two categories and the label setting is $0$ and $1$. Thus, we select label $0$ and $1$ as graph anomaly label respectively to evaluate the performance of GLADST.

\subsection{Baselines}
\begin{table*}[h!]
\caption{Anomaly detection performance measured mean value of AUC (\%) and standard deviation (\%) when graph data of label $1$ is graph anomaly.}
\centering
\begin{tabular}{ccccccc}
\hline
Datasets & FGSD-IF             & FGSD-LOF   & FGSD-OCSVM & GLocalKD   & GOOD-D     & GLADST              \\ \hline
HSE      & \textbf{60.62\tiny±1.35} & 56.56\tiny±2.49 & 57.76\tiny±4.43 & 40.92\tiny±0.98 & 54.83\tiny±3.32 & 55.47\tiny±3.23          \\
MMP      & 32.22\tiny±0.90          & 43.00\tiny±1.94 & 47.86\tiny±2.97 & 68.11\tiny±0.80 & 52.38\tiny±4.72 & \textbf{68.55\tiny±1.80} \\
P53      & 33.06\tiny±3.67          & 43.45\tiny±3.55 & 51.37\tiny±2.76 & 66.98\tiny±3.32 & 59.13\tiny±4.82 & \textbf{69.61\tiny±3.61} \\
PPAR     & \textbf{65.51\tiny±4.03} & 53.59\tiny±4.81 & 49.55\tiny±3.86 & 34.69\tiny±4.06 & 57.03\tiny±2.87 & 61.32\tiny±2.88          \\ \hline
AIDS     & 0.62\tiny±0.89           & 12.27\tiny±4.89 & 13.80\tiny±4.22 & 95.10\tiny±1.87 & 14.28\tiny±7.77 & \textbf{97.67\tiny±0.81} \\
BZR      & 55.49\tiny±6.06          & 50.44\tiny±8.73 & 58.85\tiny±6.44 & 62.57\tiny±7.32 & 29.92\tiny±9.58 & \textbf{81.02\tiny±3.00} \\
COX2     & 43.51\tiny±3.45          & 43.29\tiny±4.85 & 45.77\tiny±5.81 & 62.21\tiny±5.35 & 42.13\tiny±1.45 & \textbf{63.05\tiny±9.59} \\
DHFR     & 48.38\tiny±5.25          & 50.80\tiny±5.94 & 44.11\tiny±4.45 & 55.05\tiny±3.58 & 61.61\tiny±4.84 & \textbf{77.36\tiny±3.49} \\
NCI1     & 66.81\tiny±1.59          & 46.07\tiny±2.18 & 49.81\tiny±2.58 & 31.77\tiny±1.53 & 34.26\tiny±2.36 & \textbf{68.12\tiny±1.60} \\ \hline
ENZYMES  & 47.98\tiny±3.20          & 45.21\tiny±5.50 & 56.05\tiny±7.08 & 47.91\tiny±6.17 & 54.22\tiny±4.96 & \textbf{69.43\tiny±9.14} \\
PROTEINS & 24.60\tiny±2.79          & 40.21\tiny±3.64 & 66.36\tiny±1.64 & 56.17\tiny±3.46 & 72.35\tiny±3.34 & \textbf{78.91\tiny±3.28} \\ \hline
COLLAB   & 65.28\tiny±1.50          & 45.17\tiny±1.68 & 75.51\tiny±1.48 & 67.42\tiny±2.06 & 50.46\tiny±2.84 & \textbf{77.65\tiny±6.33} \\ \hline
\end{tabular}
\label{tab2}
\end{table*}

In the field of graph-level anomaly detection, few effective methods are put into use. Therefore, we perform the experiment with representatives from both the recent methods and the traditional methods. Firstly, we choose GLocalKD \cite{ma2022deep} as one of the baselines, which is a new deep learning method to detect graph-level anomaly. GLocalKD is capable of devising graph representations and is able to detect both local-anomaly and global-anomaly graphs better, owing to the usage of joint random distillation. In addition, some traditional methods are chosen for comparison. We select FGSD \cite{verma2017hunt} as the model for graph representation learning. Then, it is used to drive the certain anomaly detection algorithm for GLAD, including isolation forest (IF) \cite{liu2008isolation}, local outlier factor (LOF) \cite{breunig2000lof} and one-class support vector machine (OCSVM) \cite{scholkopf1999support}, hence the FGSD-IF, FGSD-LOF, and FGSD-OCSVM are included in our baselines. Furthermore, we choose a unique and recently published model, GOOD-D \cite{liu2023good}, which is an unsupervised graph out-of-distribution detection method based on contrastive learning.
\subsection{Parameter Settings}
Three models are used in the GLADST experiments, one teacher model and two student models. The identical graph encoder is applied, which is made up of double GCN layers, whose dimensions are $d$-512-256, where $d$ denotes the attribute features' dimension size in the datasets for training. For GLocalKD, we choose the recommended default parameters. We use different algorithms (IF, LOF and OCSVM) to drive FGSD while choosing the same default parameters. The paper which proposes GOOD-D provides unique parameters for each dataset, and we use them in the experiment.

\subsection{Anomaly Detection Performance}
To prove that our model performs well in many cases, we evaluate the performance of our model through comparing it with the baselines on all the aforementioned datasets. We employ 5-fold cross-validation to train these approaches and record the average AUC results along with their standard deviation. The evaluation metric then becomes the criteria by which we judge the models' effectiveness according to the previous graph anomaly detection works \cite{luo2022comga,ma2022deep}. Furthermore, to determine the influence of the selection of abnormal labels, we use different signs of graph anomaly to examine the models. In a word, we set graph data of label $0$ as abnormal graphs, and graph data of label $1$ is normal graphs, otherwise.

The AUC scores of GLADST and baselines are shown in Table \ref{tab1} and Table \ref{tab2}, respectively. We use label $1$ or $0$ as a sign of graph anomaly to observe the degree of its influence on all models. Based on experimental results presented in Table \ref{tab1}, it is obvious that the AUC results of GLADST are much better than those of the baselines most of the time, except for several datasets. Our model only obtains lower scores on HSE, MMP, PPAR, AIDS and COLLAB, and the gap between our model and the highest-scoring model is small. From the Table \ref{tab2}, GLADST outperforms all baselines apart from HSE and PPAR. Besides, the improvement of anomaly detection performance is obvious on p53, BZR, DHFR, ENZYMES and COLLAB. 

In addition, it is obvious that our model is less susceptible to interference from the selection of graph anomaly label. This is due to the symmetry of our model, which means that we have one student model trained with normal graphs and another trained with abnormal graphs. In contrast, the influence is much greater for the baselines, especially FGSD.

\subsection{Ablation Study}
\begin{figure*}[b!]
\centering

\includegraphics[width=3.6cm]{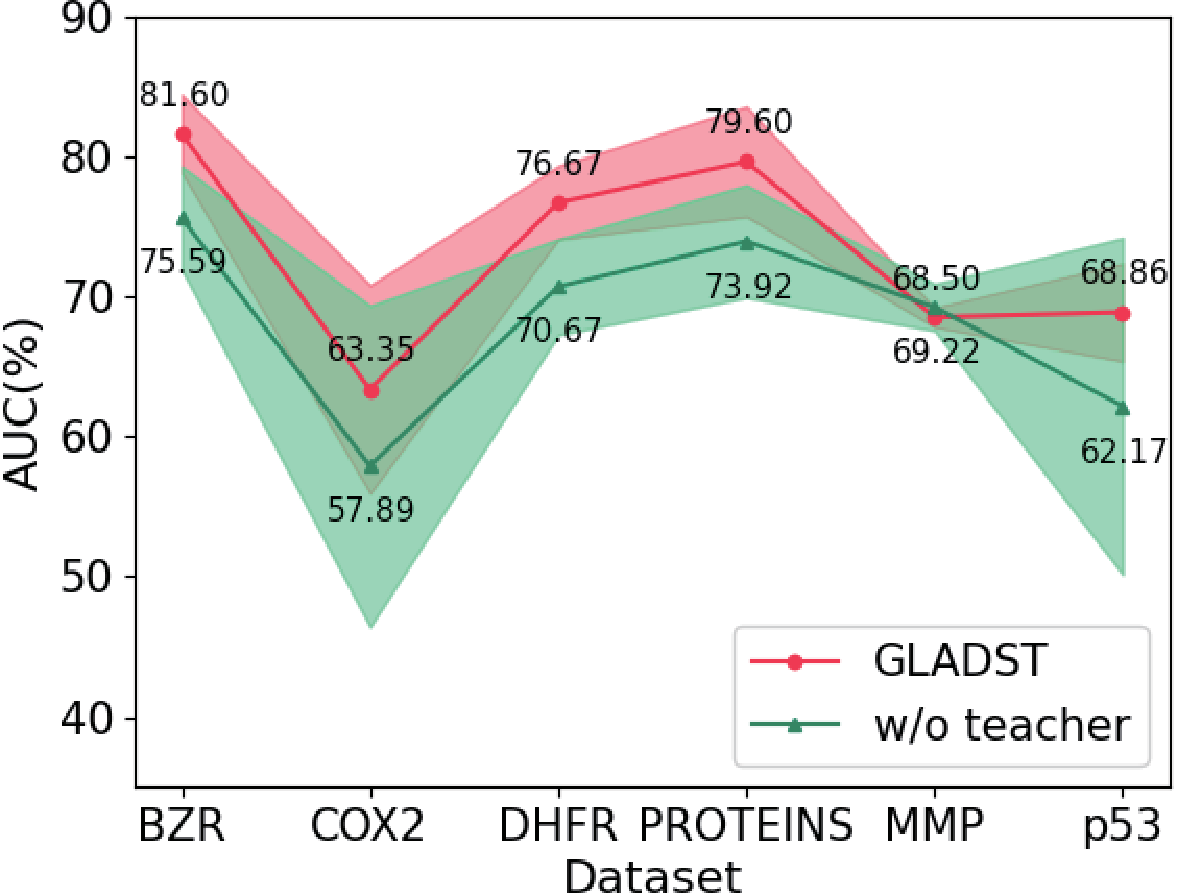} 
\quad
\includegraphics[width=3.6cm]{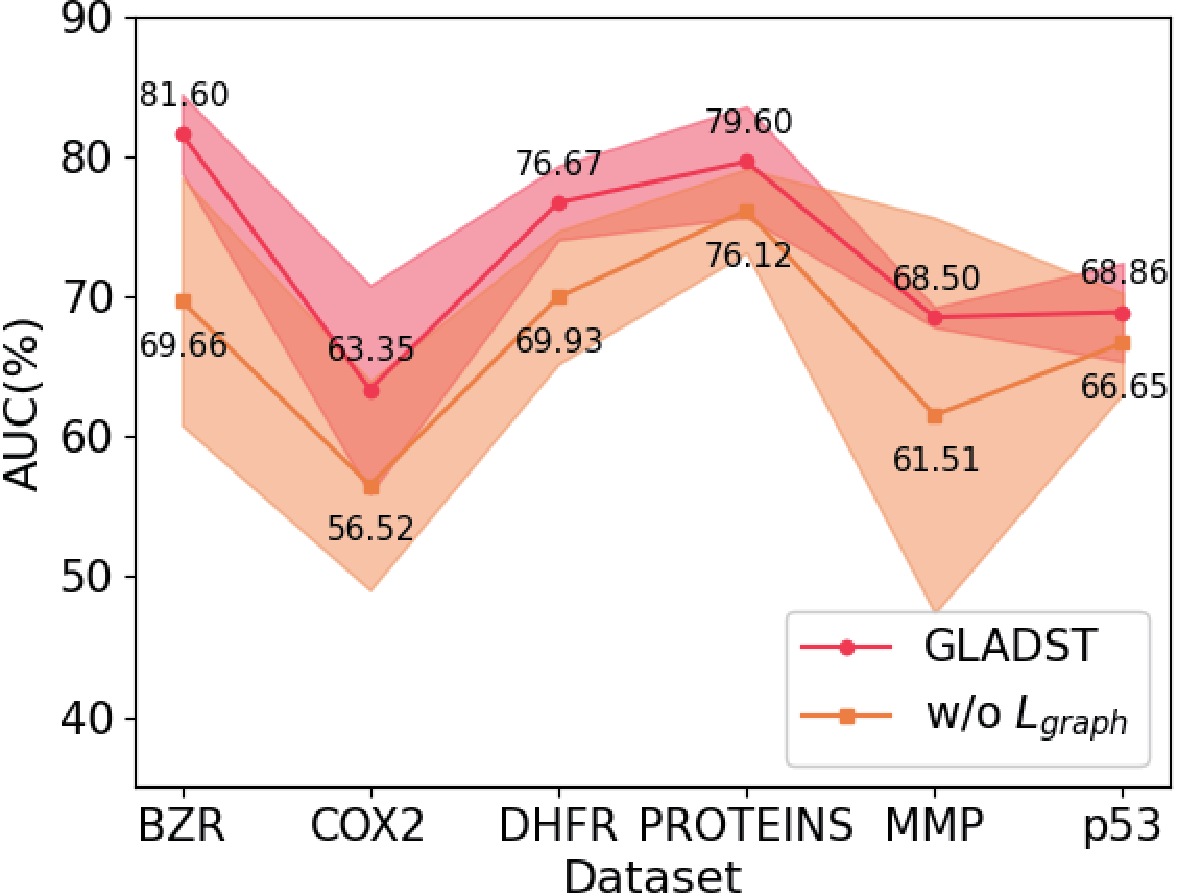}
\quad
\includegraphics[width=3.6cm]{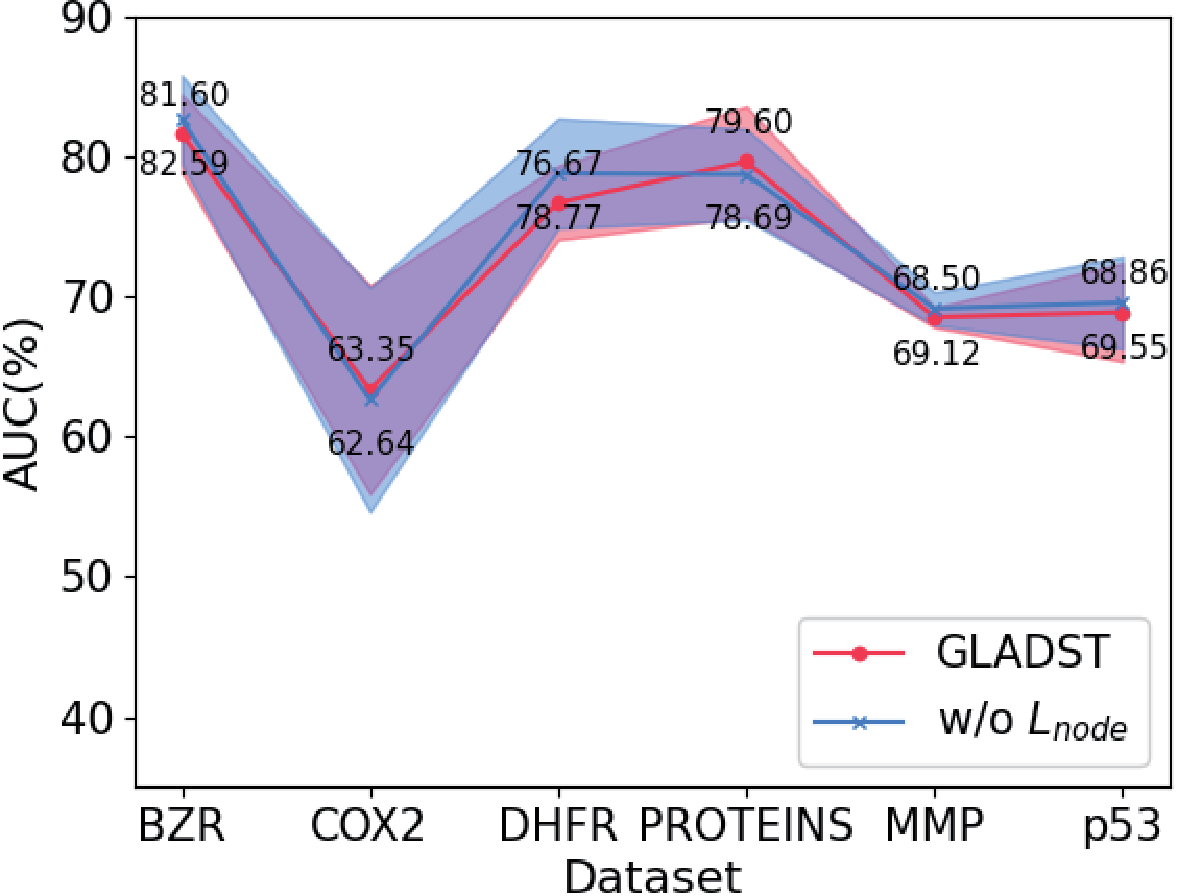}
\caption{The AUC results of model variants. The lines represent the AUC values of each model, and the shadows show the standard division.}    
\label{fig2}       
\end{figure*}
We also perform an ablation experiment on GLADST, focusing on the importance of the teacher model, node-level representation error loss ($L_{node}$), and graph-level representation error loss ($L_{graph}$). Therefore, we remove each part separately to observe their effect. Firstly, we train the student models with an untrained teacher model; then, we remove the node-level loss and graph-level loss of GLADST respectively. During the experiment, we record all the average scores and standard deviations of these models. To arrive at a high-confidence conclusion, we choose six datasets (BZE, COX2, DHFR, PROTEINS, MMP, and p53) for examination, and the sign of graph anomaly is label 0. To show our results more clearly, we present a series of graphs in Figure \ref{fig2} showing the rating scores and standard deviations for reference.

From Figure \ref{fig2}, the GLADST model shows a significant improvement when compared to the model with an untrained teacher on most of the datasets.Furthermore, when we remove the graph-level loss in the model, the scores on most datasets decrease dramatically. But the model without node-level loss seems to be only a little affected. Overall, our model makes progress in terms of performance. The results demonstrate that the design of the node-level and graph-level losses are effective to achieve node property and graph property anomaly detection. The results also show that significant improvement is made due to the trained teacher which contributes to making a more obvious distinction between the feature from the teacher and the feature from the student model without training.

\begin{figure}[b!]
\centering
\subfigure[BZR]{ 
\includegraphics[width=3.6cm]{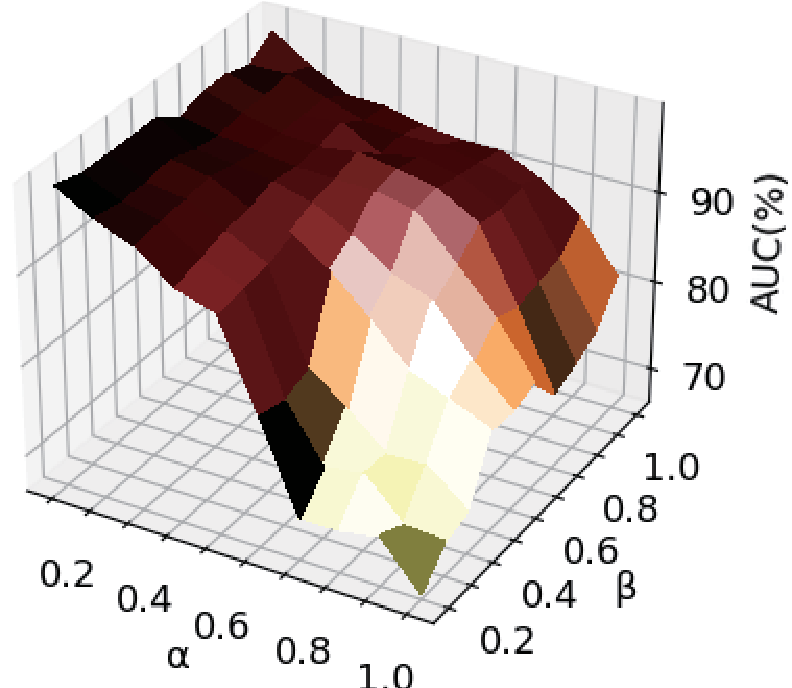} 
}\subfigure[COX2]{ 
\includegraphics[width=3.6cm]{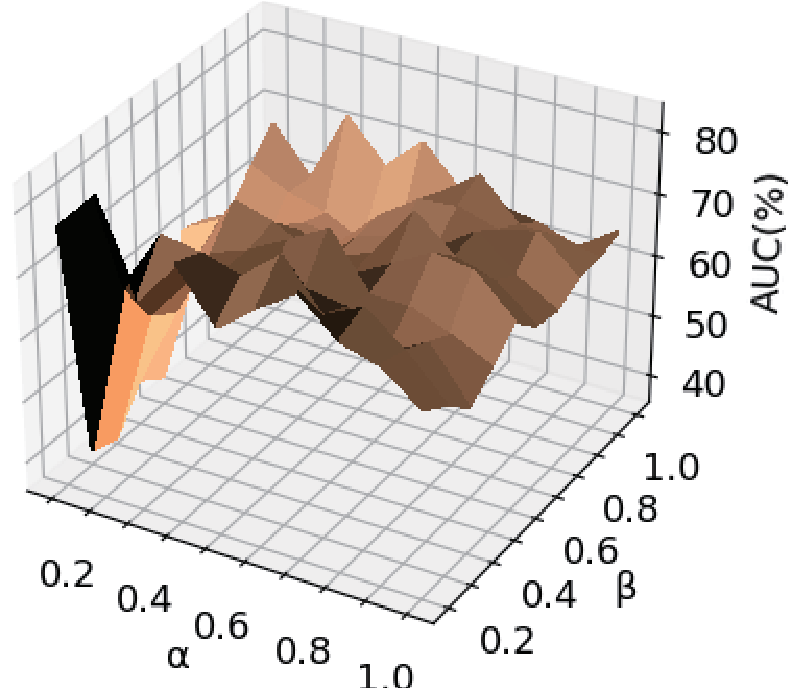}
}\subfigure[DHFR]{  
\includegraphics[width=3.6cm]{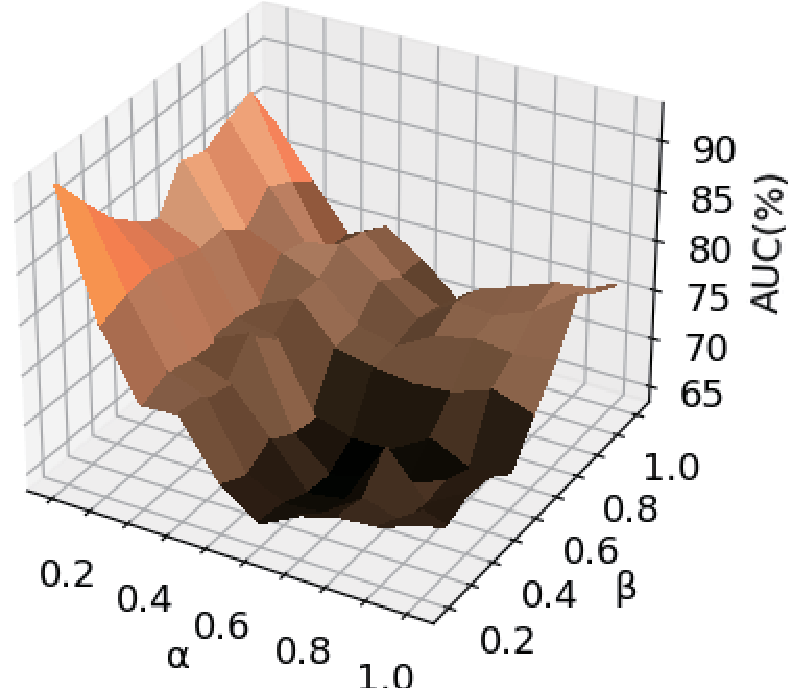}
}
\subfigure[PROTEINS]{  
\includegraphics[width=3.6cm]{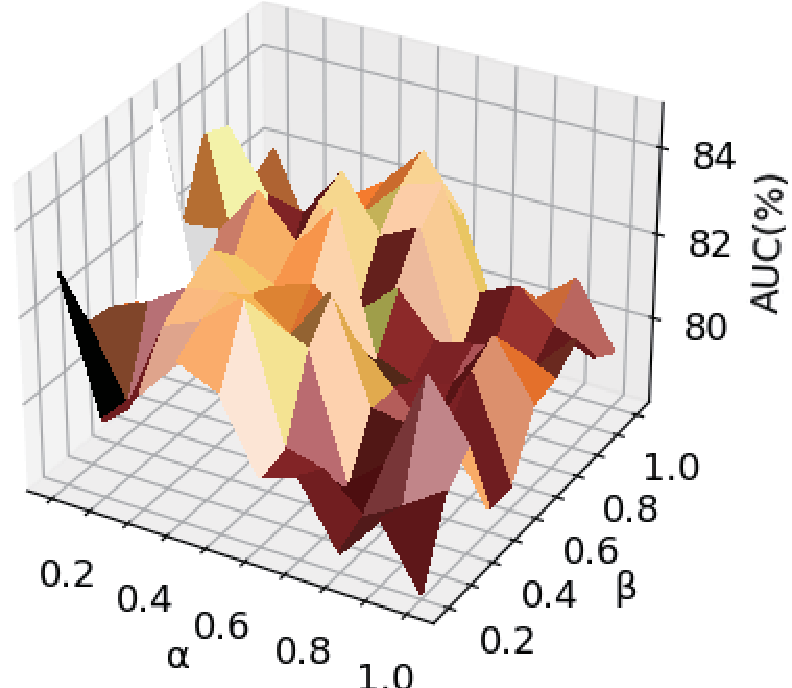}
}\subfigure[MMP]{  
\includegraphics[width=3.6cm]{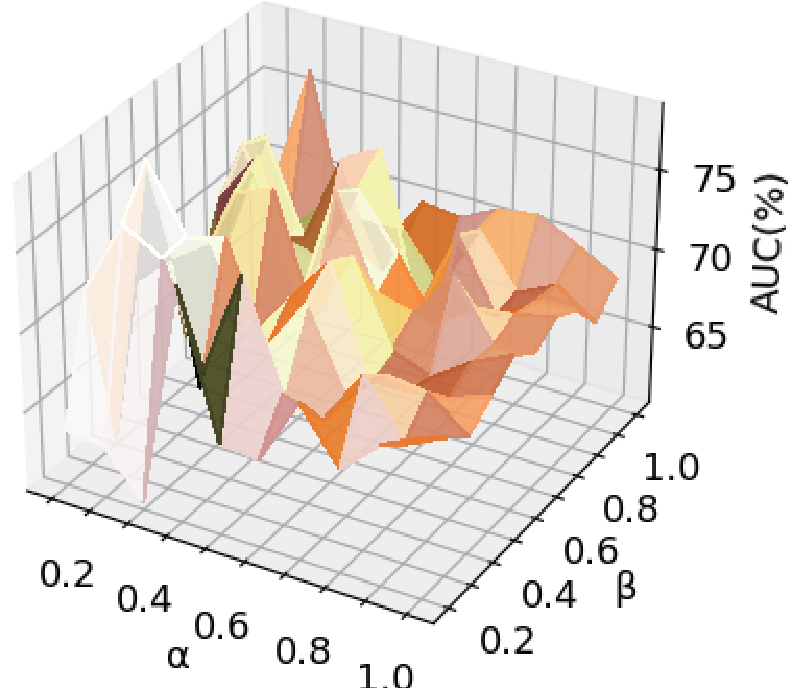}
}\subfigure[p53]{  
\includegraphics[width=3.6cm]{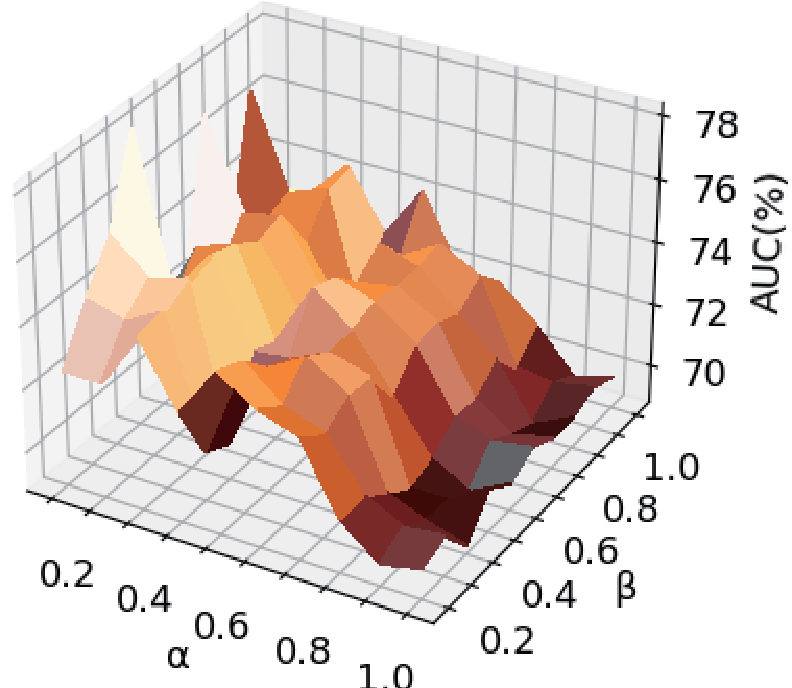}
}
\caption{The AUC results of GLADST under a different number of anomalies in the test set and training set. $\alpha$ represents the proportion of the anomalies to the original total anomalies of the test set. Similarly, $\beta$ represents this in the training set.}   
\label{fig3}       
\end{figure}
\subsection{Efficiency Analysis}
In this section, we mainly explore the impact of varying the number of abnormal samples in both the test and training sets. To begin with, we choose label 0 as the sign of graph anomaly and the same datasets in our ablation study. We next divide each dataset into a test set and a training set using the ratio of four to one. Then, we separate the anomalies from both sets. After this, we add 10\% of the anomalies into the training set each time to train the model and do the same thing with the test set. Our results are shown in Figure \ref{fig3}.

Figure \ref{fig3} illustrates that with an increase in anomalies in the training set, the AUC results fluctuate and improve to a certain extent when the anomalies in the testing set stay the same. With an increase in abnormal samples in the test set, the results of AUC decrease sharply when the training set is invariant. The overall AUC results change to be within a stable and acceptable range. Furthermore, it is obvious that even when the quantity of anomalies in the training set or test set is not large, GLADST remains valid.

\subsection{Visualization Analysis}
\begin{figure}[b!]
	\small
	\includegraphics[width=\linewidth]{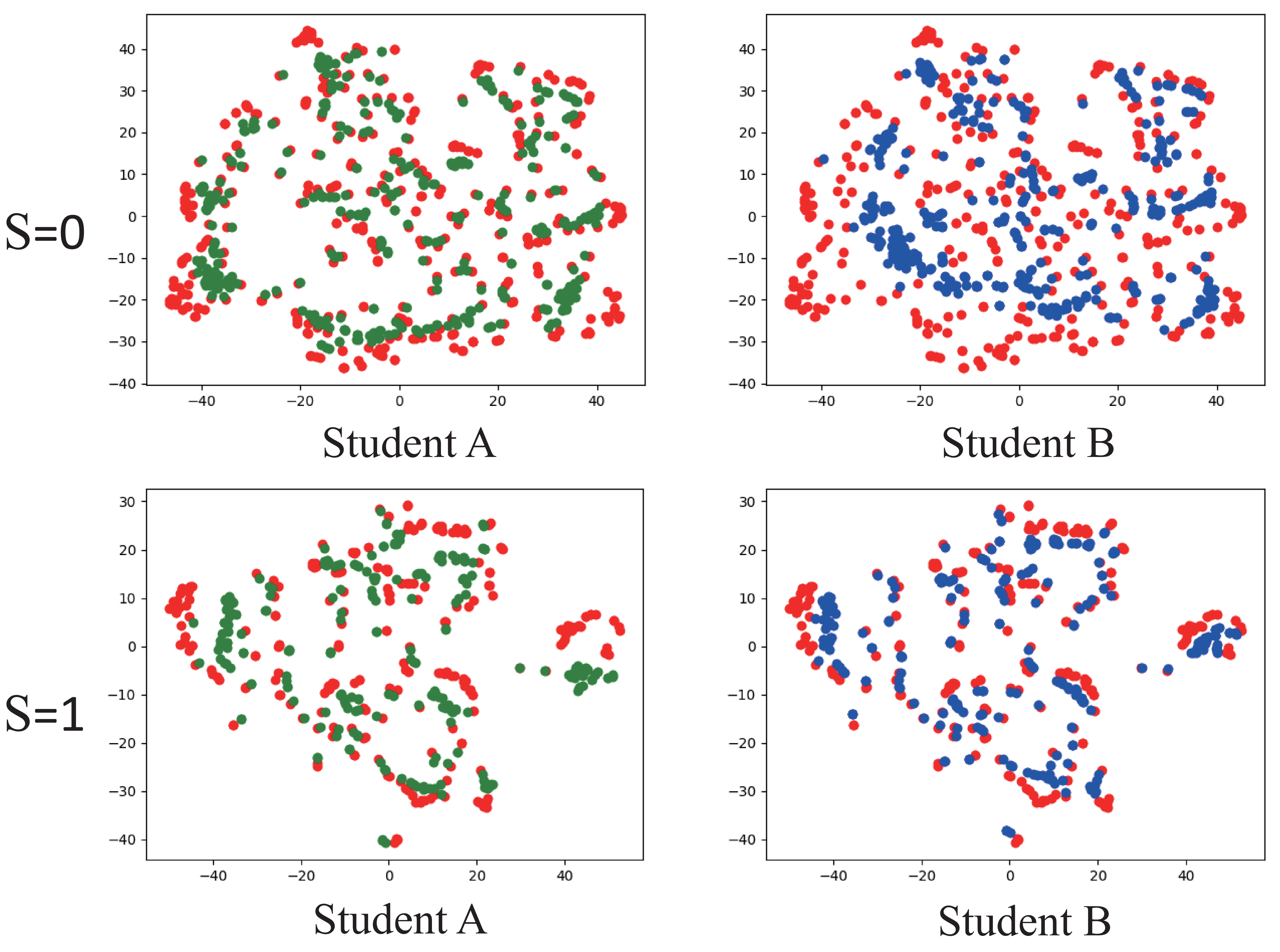}
	\caption{A visualization of GLADST performance on the DHFR dataset, where red denotes the graph feature representations of the trained teacher model, and green and blue denote the graph feature representations for student A with normal graphs training and student B with abnormal graphs training, respectively. $S=0$ and $S=1$ denote the inputs of models with normal graphs and abnormal graphs, respectively.} \label{fig6}
\end{figure}
For an anomalous graph, we consider node property and graph property anomalies in the graph and apply node-level representation and graph-level representation error losses to achieve anomaly detection in the GLAD task. To intuitively represent the effect of the proposed GLADST for anomalous graph detection, we first train our method based on the dual-students-teacher model on the training set for the DHFR dataset and then give the test graphs to evaluate its efficiency. We visualize the experiment results in Figure \ref{fig6}. Thus, we can see that when we input the normal graphs $S=0$ into student A and student B, respectively, the feature representation of student A is closer to the feature representation of the teacher model than student B. That is to say, the joint error loss $\hat{S}_{G}$ of student A is smaller and the $\check{S}_{G}$ of student B is larger, so the difference between them as an anomaly score will be less than $0$, which is judged as a normal graph. Similarly, for the abnormal graphs $S=1$, the anomaly score will be greater than $0$, which is judged as an abnormal graph. The practical analysis above demonstrates the effectiveness of utilizing node-level and graph-level representation error losses to perform anomalous graph detection and designing a dual-students-teacher model to train the GLAD framework. 
\section{Conclusion}
We explore the key problem that node property and graph property anomalies are very important to anomalous graph detection. To design a powerful evaluation mechanism to distinguish anomalous graphs, we introduce a discriminative graph-level anomaly detection framework via dual-students-teacher model. Through the optimization of node-level and graph-level representation error losses between two student models and a trained teacher model, respectively, the value of representation error between two student models as the score function can be effectively detected anomalous graphs. The outstanding performance of GLADST compared with five baselines on twelve real-life datasets demonstrates the effectiveness of our method. Furthermore, the ablation study, the efficiency analysis, and the visualization experiments also verify that our model design considerably improves the graph-level anomaly detection performance.

\subsubsection{Acknowledgements}  The research work is supported by Wuhan University People's Hospital Cross-Innovation Talent Project Foundation under JCRCZN-2022-008. 

\bibliographystyle{splncs04}
\bibliography{sample}
\end{document}